%% file: acl_latex.tex
\documentclass[11pt]{article}

\usepackage[final]{acl}

\usepackage{microtype}

\usepackage[english,bidi=default]{babel} % English as the main language.

\newcommand*{\ENDASH}{\babelhyphen{--}}
\newcommand*{\FOO}{\babelhyphen{hard}}

\babelfont{rm}{TeXGyreTermesX} % similar to Times
\babelprovide[import]{farsi}
\babelfont[*arabic]{rm}[
Path=./fonts/ttf/,
Scale=0.9,
Extension = .ttf,
UprightFont=*-Regular,
BoldFont=*-Bold,
]{Vazirmatn}

\usepackage{graphicx}
\usepackage{booktabs}
\usepackage{amsmath}
\usepackage{amssymb}

\usepackage[normalem]{ulem}
\usepackage{fontawesome5}
\usepackage{tikz}
\usetikzlibrary{arrows.meta,shapes.geometric,calc,positioning,backgrounds}

\usepackage{mathtools}

\usepackage{listings}
\usepackage{xcolor}

\usepackage{tcolorbox}
\tcbuselibrary{breakable}

\newcommand{\FactsPers}{X_p}
\newcommand{\FactsEn}{X_e}
\newcommand{\ShallowCues}{M}
\newcommand{\Prediction}{Y}
\newcommand{\cace}{\mathrm{ACE}^{\star}}
\newcommand{\E}{\mathrm{E}}
\newcommand{\causalDo}{\mathrm{do}}

\newcommand{\semrel}[1]{\textsc{#1}}
\newcommand{\triple}[3]{$\langle$#1, \semrel{#2}, #3$\rangle$}

\title{Probing Factual Knowledge Transfer with Training Data Interventions}

\author{Romina Oji \quad Marc Braun \quad Marcel Bollmann \quad Marco Kuhlmann \quad Jenny Kunz \\
  Department of Computer and Information Science \\
  Linköping University \\
  \texttt{firstname.lastname@liu.se} \\}

\begin{document}
\maketitle
\begin{abstract}
Do multilingual language models transfer factual knowledge across languages during continued pretraining, or do they mostly recall facts learned directly from the target-language data? To answer this question more reliably, we propose an intervention-based framework: starting from an English-pretrained model, we continue pretraining on Persian data from which specific facts have been systematically removed at varying levels of granularity. We construct SIFT, a resource of 500 triples across 20 relations, stratified by the cultural origin of each fact's subject into general (globally prominent) and Persian{\FOO}related entities, designed for both systematic fact removal from training data and evaluation, with natively written Persian cloze templates. 
Our results show that fact transfer is very limited: under the strictest removal condition, a large majority of English-acquired facts fail to transfer into Persian. We further show that sentence-level co-occurrence removal is insufficient to eliminate fact signal, and that easier (randomly selected) negative candidate sets substantially inflate apparent transfer by rewarding shallow associative heuristics, while performance on a harder candidate set that allows for less reliance on heuristics is much lower. Finally, we show that source-language entity frequency has a large influence, with Persian{\FOO}related facts, which are orders of magnitude rarer in the English corpus, hardly transferring. 
\end{abstract}

\section{Introduction}

Multilingual language models are able to retrieve factual knowledge acquired in one language when prompted in another \cite{petroni-etal-2019-language, kassner-etal-2021-multilingual}, but cross{\FOO}lingual fact transfer can be inconsistent \cite{qi-etal-2023-cross, wang-etal-2025-lost-multilinguality}.
Several studies have attempted to characterize which facts transfer across languages and which do not \cite{jiang-etal-2020-x, keleg-magdy-2023-dlama}; however, it is hard to draw general conclusions without full access to the training corpus.

\begin{figure}
    \resizebox{\linewidth}{!}{%
        \input{marco}
    }
    \caption{Overview of our causal framework. Our bilingual English{\ENDASH}Persian language model could recall facts in Persian because (a)~it has acquired the knowledge from the English corpus and transferred it; (b)~it has directly acquired the knowledge during continued pretraining on Persian; (c)~it has derived the correct answer from shallow cues. Our interventions are designed to block paths~(b) and~(c), thereby quantifying the effect size of cross{\FOO}lingual transfer,~(a).}
    \label{fig:schematic}
\end{figure}
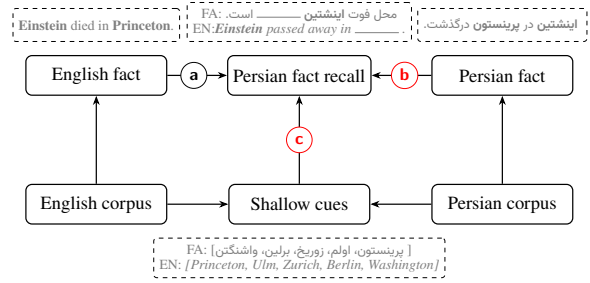

In this paper, we focus on a more tractable scenario of cross{\FOO}lingual transfer:
We build bilingual language models through continued pre{\FOO}training of an English base model on Persian data.
This setup suggests a concrete causal model of how fact transfer could occur (see\ Figure~\ref{fig:schematic}) and enough experimental control to test whether it does.
More specifically, our bilingual English{\ENDASH}Persian models could recall facts in Persian because they have either (a)~transferred them from the English training corpus or (b)~acquired them during the continued pretraining on the Persian corpus.
Inspired by recent work on the role of heuristics in fact recall \citep{elazar2022measuring,saynova-etal-2025-fact}, we also allow for a third causal path (c), where the model could derive the correct answer from shallow cues such as geographic or associative co{\FOO}occurrence.
Our experiments are designed around interventions to approximately block paths (b) and (c), which lets us estimate how much of the model's performance can be causally attributed to actual knowledge transfer \citep{pearl2009causality}.
In particular, we intervene on~(b) by removing facts from the Persian corpus in controlled ways, and on~(c) by increasing the difficulty of fact retrieval through strong distractors that reduce the usefulness of shallow cues.
As the key resource for both interventions, we construct SIFT (Stratified Intervention Facts for Transfer), a resource of 500 fact triples across 20 relations, annotated by the provenance of the subject entity: general facts, whose subjects are widely recognized around the world, and Persian{\FOO}related facts, whose subjects are culturally rooted in the Persian{\FOO}speaking world.

% As the key resource for both interventions, we construct SIFT (Stratified Intervention Facts for Transfer), a resource of 500 fact triples across 20 relations and annotated with two categories of provenance (general vs.\ Persian{\FOO}related facts).

We use the English $\to$ Persian scenario as a challenging case study for cross{\FOO}lingual transfer.
The Persian script shares no orthographic overlap with English and the two languages have virtually no common vocabulary, requiring the model to build cross{\FOO}lingual associations entirely through learned semantic representations.
Indeed, under our strictest conditions, where we intervene on the training data by removing all mentions of the facts' subjects from the Persian corpus and evaluate with hard distractors that suppress shallow associative cues, we find very limited signs of cross{\FOO}lingual fact transfer, consistent with prior findings that even large-scale models struggle to share factual knowledge across languages \citep{goldman2025eclektic, xu-etal-2023-language-representation, gao-etal-2024-multilingual}.

The model transfers only a small fraction of its English{\FOO}acquired knowledge into Persian, with transfer largely confined to general facts, while Persian{\FOO}related facts transfer at substantially lower rates. To understand this gap, we investigate the relationship between how frequently a fact's subject appears in the English pretraining corpus and whether that fact transfers to Persian. Persian{\FOO}related subjects, which are orders of magnitude rarer in English, transfer at near-chance levels, whereas highly frequent Persian{\FOO}related subjects reach moderate transfer rates. These findings establish that a strong representation in the English data is necessary but not sufficient for fact transfer.

\section{Related Work} \label{sec:related-work}

LAMA \citep{petroni-etal-2019-language} introduced cloze{\FOO}style probing of factual knowledge over subject{\FOO}relation{\FOO}object triples, and subsequent work extended this paradigm to multilingual and cross{\FOO}lingual settings \citep{kassner-etal-2021-multilingual, jiang-etal-2020-x, elazar-etal-2021-measuring, fierro-sogaard-2022-factual, qi-etal-2023-cross, wang-etal-2025-lost-multilinguality, liu-etal-2025-tracing, keleg-magdy-2023-dlama}.
% The dataset most related to ours is ECLeKTic \citep{goldman2025eclektic}, which also aims to evaluate cross-lingual factual transfer by selecting facts that appear in Wikipedia in only one language, under the assumption that such facts are present in the training data of that language alone, allowing them to test whether models can retrieve these facts when queried in another language. 
However, all of these datasets are designed solely for evaluation; none provides facts structured for systematic removal from training data, which our study requires to isolate transfer from direct learning.

Cross{\FOO}lingual factual knowledge sharing in multilingual models remains limited: high- and low{\FOO}resource languages show significant performance gaps in fact probing \citep{xu-etal-2023-language-representation}, multilingual pretraining and instruction tuning yield only shallow knowledge alignment \citep{gao-etal-2024-multilingual}, and even state{\FOO}of{\FOO}the{\FOO}art models struggle to retrieve facts when queried in a language other than the one in which the fact is expressed \citep{goldman2025eclektic}.
Furthermore, models may not genuinely store facts but instead rely on statistical shortcuts such as prompt biases, co-occurrence statistics, and dataset artifacts \citep{cao-etal-2021-knowledgeable, elazar2022measuring}, with evidence that genuine recall, heuristic shortcuts, and guesswork activate distinct model mechanisms \citep{saynova-etal-2025-fact}.
Fact recall also scales with entity frequency in the pretraining data, with long{\FOO}tail entities remaining difficult even at large model scales \citep{kandpal2023large}.
These findings motivate our use of hard and random candidate sets to disentangle precise knowledge from shallow cues, and our analysis of subject frequency as a predictor of cross-lingual transfer (\S\ref{subsec:freq-transfer}).

\section{SIFT}

To support our analysis of cross{\FOO}lingual fact transfer, we introduce SIFT\footnote{The dataset is available on
\href{https://huggingface.co/datasets/liu-nlp/SIFT}{Hugging Face}.} (\textbf{S}tratified \textbf{I}ntervention \textbf{F}acts for \textbf{T}ransfer), a resource of 500 factual knowledge triples annotated for subject provenance (general vs.\ Persian{\FOO}related), natively written Persian templates, and negative candidate sets (distractors) at two difficulty levels, designed for both training data intervention and evaluation.

As discussed in \S\ref{sec:related-work}, existing fact probing datasets were designed for evaluating fact recall, not for the controlled manipulation of training data that our study requires. LAMA-style datasets have several shortcomings for our purposes: Many triples have minimal coverage in the Persian web corpus, making their removal from training data an empty exercise rather than a meaningful intervention. Some triples are trivially self{\FOO}referential, such as \triple{flag of Sweden}{applies\_to\_jurisdiction}{Sweden}, and are unlikely to appear as natural text and too easy to serve as meaningful evaluation items. Also, their facts are selected from English{\FOO}centric sources \citep{kassner-etal-2021-multilingual, fierro-sogaard-2022-factual} with no coverage of Persian{\FOO}related entities. ECLeKTic \citep{goldman2025eclektic} addresses cross-lingual transfer more directly, but by selecting facts available in only one Wikipedia language, it inherently restricts evaluation to culture-specific or rare entities, excluding the general facts that are essential for a complete picture of cross-lingual transfer. SIFT addresses these gaps by selecting triples with sufficient Persian corpus frequency to make removal meaningful, while avoiding entities so frequent that their removal would distort the training data, and includes Persian{\FOO}related entities absent from existing English-centric resources.

\subsection{Fact Triples}\label{subsec:factual_triples}

We represent each fact as a subject--relation--object triple \triple{s}{r}{o}, a representation widely used in factual knowledge probing \citep{petroni-etal-2019-language, wang-etal-2025-lost-multilinguality}. For example, the fact that Albert Einstein died in Princeton is represented as \triple{Albert Einstein}{place\_of\_death}{Princeton}. In our study, these triples serve a dual purpose: they define the targets for our training data intervention,
%: we use the triples to identify and remove sentences containing the corresponding fact from the target-language corpus.
and they form the basis of our evaluation of factual knowledge recall. Relations are language-independent Wikidata-style identifiers, while subjects and objects are stored with surface forms in both English and Persian: the English forms are used for English evaluation, and the Persian forms, which include all orthographic and lexical variants of an entity (\S\ref{subsec:removal}, Appendix ~\ref{app:triples-criteria}), are used for the training-data intervention and for Persian evaluation. In addition to the triple itself, each entry in SIFT carries corpus statistics for the subject: its frequency in the English and Persian pretraining corpora, and the number of sentences in which the subject and object co-occur. We use these statistics to select triples whose removal is a meaningful intervention (Appendix~\ref{app:triples-criteria}) and to analyse how source-language frequency affects transferability (\S\ref{subsec:freq-transfer}).
%: given a cloze template constructed from the subject and relation, the model is expected to predict the object.

We construct a set of fact triples in two categories based on the provenance of the subject:
(a) \emph{Persian{\FOO}related facts}, whose subjects are culturally rooted in Iran or the Persian-speaking world,
% , such as an Iranian musician and the traditional instrument they play;
% these facts appear in English sources, but the entities themselves belong to Persian cultural context;
and (b) \emph{general facts}, whose subjects are globally prominent with no strong tie to any particular region. Prior work has shown that multilingual models systematically favor Western entities and store culturally specific knowledge less reliably than globally shared knowledge \citep{hershcovich-etal-2022-challenges, naous-etal-2024-beer, yin-etal-2022-geomlama}. Our categorization allows us to analyze how the provenance of a fact affects its cross-lingual transfer.

We construct our evaluation set through a combination of LLM-assisted generation and manual curation. We prompted Claude Opus 4.6 with extended thinking via the Claude chat interface (claude.ai) in February--March 2026 to generate 70 candidate triples per relation across the two provenance categories. A native Persian speaker manually reviewed all candidates against Wikidata and Wikipedia, discarding or replacing triples that were factually incorrect, ambiguous, or insufficiently attested in the Persian web corpus. Persian{\FOO}related triples required substantial manual intervention, as the model repeatedly generated the same small set of subjects across relations, limiting diversity. The final dataset comprises 500 triples\footnote{Since the same subject may appear in multiple relations, the number of unique subjects (437) is smaller than the total number of triples. This also holds within each provenance category: 324 unique general subjects vs.\ 386 general triples, and 113 unique Persian-related subjects vs.\ 114 Persian-related triples (Table~\ref{tab:frequency-by-culture}).} across 20 relations (25 per relation), annotated as Persian{\FOO}related~(114) or general~(386). The distribution of Persian{\FOO}related and general triples is not uniform across relations, as some relations, such as \semrel{capital} and \semrel{continent}, are inherently general, while person-centric relations can include both general and Persian{\FOO}related subjects. Additional curation criteria are described in Appendix~\ref{app:triples-criteria}.

\subsection{Evaluation Templates}

For each relation, a template is a declarative sentence containing a placeholder for the subject and a placeholder for the object, written such that substituting the correct object yields a natural, factually true sentence. At evaluation time, we fill the subject slot with the fixed subject of a triple and score each candidate object by substituting it into the object slot and computing its length-normalized log-probability under the model (\S\ref{subsec:evaluation}); the candidate with the highest score is taken as the model's prediction. For example, as illustrated in Figure~\ref{fig:schematic} a template for the relation \semrel{place\_of\_death} can be ``\{subject\} passed away in \{object\}'' which, for the triple \triple{Albert Einstein}{place\_of\_death}{Princeton}, would be scored once with the correct object (``Albert Einstein passed away in Princeton'') and once for each of the four incorrect object candidates (e.g., ``Albert Einstein passed away in Berlin''). We show the English template here for readability; the Persian templates are written natively and are not translations of their English counterparts.

Existing multilingual probing datasets rely on machine-translated templates, which introduces grammatical and semantic artifacts in the target language \citep{semenov-sennrich-2025-measuring, kassner-etal-2021-multilingual, fierro-sogaard-2022-factual, wang-etal-2025-lost-multilinguality}. We therefore write new declarative templates natively in Persian, where the object can appear at any position in the sentence, better approximating naturally occurring text in the pretraining corpus. For each relation, we create five templates, all written and reviewed by a native Persian speaker.\footnote{We additionally create question--answer format templates for comparison with prior work; results are reported in Appendix \ref{app:mean-results}.} For English evaluation, we use all five English templates of KLAR \citep{wang-etal-2025-lost-multilinguality} per relation; these are also written manually.

\subsection{Candidate Sets}
To evaluate whether models can distinguish the correct object from plausible alternatives, we construct four negative candidate objects for each triple, in addition to the correct one. Candidates are negative in the sense that substituting them for the correct object yields a triple representing a provably false ``fact.'' Each candidate is itself a real entity of the same type, verified against Wikidata and Wikipedia to be incorrect for the given subject and relation. We construct candidates in two difficulty settings.

In the \emph{hard} setting, candidates are designed to be semantically close to the correct answer. For each relation, we defined criteria guiding the generation of plausible but incorrect candidates. For example, candidates for \semrel{capital} include the country's largest non-capital city, neighboring country capitals, and cities with similar spelling; candidates for \semrel{place\_of\_birth} and \semrel{place\_of\_death} include cities from the cultural or geographic region suggested by the person's name, so candidates for an Iranian figure are other major Iranian cities rather than unrelated world capitals. We prompted Claude Opus 4.6 with extended thinking via the Claude chat interface in February–March 2026 and these relation-specific criteria to generate candidates (full prompt in Appendix~\ref{app:candidate-prompt}), then verified that all candidates were factually incorrect and sufficiently proximate to serve as challenging distractors. In the \emph{random} setting, we sample four candidates uniformly from all objects appearing in the same relation, without regard to semantic proximity.

The hard and random settings test different aspects of fact recall. In the random setting, distractors are rarely close to the correct object, so surface-level cues, such as associating a Persian name with Iranian cities, are often sufficient to identify the correct answer without knowing the fact. Hard distractors are drawn from the same semantic and cultural neighborhood as the correct object, so these cues discriminate poorly among them, and success in the hard setting reflects a correspondingly smaller contribution from such heuristics. Assuming that performance in the hard setting approximates performance under an intervention that removes shallow cues, the hard candidate setting allows us to block path (c) in Figure \ref{fig:schematic}, better isolating the contribution of English factual knowledge to the prediction task.

\section{Experimental Setup}
% Our experimental design investigates whether factual knowledge transfers from English to Persian during continued pretraining by systematically removing specific facts from the Persian training data and testing whether models can still recall them.  We  first describe our choice of training framework (\S\ref{subsec:trainingframework}), the fact removal procedure applied to the training corpus (\S\ref{subsec:removal}), and the resulting set of models (\S\ref{subsec:model}).

Our experimental design operationalizes the causal framework in Figure \ref{fig:schematic}. We describe the training framework (\S\ref{subsec:trainingframework}),  apply three levels of fact removal intervention to the Persian training corpus, and construct one control condition (\S\ref{subsec:removal}), train models on these variants  (\S\ref{subsec:model}), and evaluate fact recall and cross{\FOO}lingual transfer using the SIFT triples and templates (\S\ref{subsec:evaluation}).

\subsection{Training Framework}\label{subsec:trainingframework}
% A key requirement of our experimental design is full transparency over the training data: we must know exactly which facts a model has been exposed to in each language. This rules out using off-the-shelf multilingual models such as LLaMA \cite{touvron2023llama} or BLOOM \cite{workshop2023bloom176bparameteropenaccessmultilingual}, whose training corpora are either undisclosed or too large to audit for specific factual content.

We adopt the training framework of \citet{glocker-etal-2025-grow}, who train language models from scratch using publicly available data with full control over the training pipeline. Their framework offers several properties well-suited to our study. First, they train a 180M-parameter seed model on English data from FineWeb-Edu \cite{lozhkov2024fineweb-edu} and code from Python-Edu \cite{benallal2024smollmcorpus}, then upscale it to 572M parameters via HyperCloning \cite{pmlr-v262-samragh24a}, a method that preserves the smaller model's output distribution while increasing capacity. This gives the upscaled model a warm start that retains its English capabilities before continued pretraining on a target language, which in our case is Persian. Second, they demonstrate that upscaled models better preserve the base model's English knowledge during continued pretraining, reducing catastrophic forgetting, a critical property for our study, since we rely on English knowledge persisting through Persian adaptation.

\subsection{Training Data Interventions}\label{subsec:removal}
To block the path from Persian training data to Persian factual knowledge recall (path (b) in Figure~\ref{fig:schematic}), we apply three levels of intervention to the Persian FineWeb-2 corpus \citep{penedo2025fineweb2pipelinescale}, each removing information about the target entities with different granularity.\footnote{Code for the data interventions and evaluation is available at \url{https://github.com/liu-nlp/SIFT}.} All levels share the same preprocessing and matching pipeline. We split the corpus into sentences using ParsiNorm \cite{oji2021parsinorm}, and for sentences exceeding 300 characters, we further split them into chunks at word boundaries while preserving complete tokens. 
% Matching requires handling two properties of Persian text. First, zero-width non-joiners (ZWNJ) are used inconsistently in Persian writing; for example, \foreignlanguage{arabic}{``آلن تورینگ''} (with space) and \foreignlanguage{arabic}{``آلن‌تورینگ''}(with ZWNJ) refer to the same entity. We normalize both the sentence and the search terms before matching, but retain the original unnormalized text when no match is found. Second, subject and object tokens are frequently concatenated with punctuation in Persian; we strip punctuation boundaries during matching to ensure detection. In all conditions, removed or replaced sentences are stored separately for analysis, and adjacent sentences are concatenated to maintain document coherence.

\paragraph{Co-occurrence removal} For each triple \triple{s}{r}{o}, we generate all pairwise combinations of subject and object surface forms -- e.g., given the triple \triple{Albert Einstein}{field\_of\_work}{physics}, this yields 2 variants of ``Einstein'' $\times$ 5 Persian variants of ``physics'' $=$ 10 pairs. We then remove any sentence in which the subject and object from at least one such pair co-occur. The need for multiple object surface forms arises because Persian allows several orthographic and lexical variants for the same concept (see Appendix~\ref{app:triples-criteria}, exhaustive object enumeration).

\paragraph{Full subject removal} We remove all sentences containing any variant of the subject, regardless of whether the object co-occurs. This provides a stronger blocking of path (b) than co-occurrence removal, as it eliminates any Persian-language context in which the entity appears, ensuring that the model cannot acquire any fact about the subject from the Persian training data.

\paragraph{Full template replacement} All sentences containing a subject are replaced with neutral template sentences that preserve the subject's surface form without exposing the model to the target factual associations. This allows us to test whether familiarity with the entity's Persian name, independent of factual context, affects the model's ability to transfer knowledge from English. Details on generation of replacement sentences are in Appendix \ref{app:replace}.

\paragraph{Random removal control} The three interventions above all reduce the size of the Persian corpus, most substantially under full subject removal. To separate the effect of removing fact-specific content from the effect of removing text, we additionally consider a control data set in which sentences are removed at random, except that every sentence containing any variant of a SIFT subject or object is retained. This corpus is reduced by an amount of text comparable to that in full subject removal while leaving all factual evidence in place. Unlike the three interventions, it is not intended to block path (b); it separates
the effect of removing facts from the effect of removing data.

\begin{table*}[h]
\resizebox{\linewidth}{!}{%
\begin{tabular}{@{}l l r r r l@{}}
\toprule
\textbf{Model} & \textbf{Training Data} & \textbf{English Tokens}& \textbf{Persian Tokens} & \textbf{Removed} & \textbf{Description} \\
\midrule
\multicolumn{5}{@{}l}{\textit{Baselines}} \\[2pt]
EN-Base  & English only           & 187.355B          & --              & --                  & No Persian exposure \\
FA-Base  & Persian only           & --          & 60.535B         & --                   & No English exposure \\
\midrule
\multicolumn{5}{@{}l}{\textit{Continued pretraining of EN-Base (+ 12\% English replay)}} \\[2pt]
CPT-Full   & Full Persian corpus          & 8.257B        & 60.535B  & --                   & Unfiltered baseline \\
CPT-CoocRM & Co-occurrences removed          & 8.257B     & 60.163B  &  0.61\%   & Sentences with subject \emph{and} object removed \\
CPT-SubjRM & All subject mentions removed          & 8.257B  & 55.392B  & 8.50\%      & Sentences with subject removed \\
CPT-SubjRP & Subject mentions replaced          & 8.257B  & 55.638B  & 8.09\%      & Sentences containing subject replaced by \\&&&&&  neutral templates to preserve surface form \\
CPT-RandRM & Random sentences removed          & 8.257B  & 55.819B & 7.79\%      & Random sentences containing no subject or \\&&&&&  object removed \\
\bottomrule
\end{tabular}
}
\caption{Overview of models and training data conditions. \textit{Removed} is relative to the full Persian corpus.
}
\label{tab:models}
\end{table*}

\subsection{Models}\label{subsec:model}
We use seven models, summarized in Table~\ref{tab:models}, that vary in their language exposure and the degree of fact removal applied to the Persian training data.\footnote{The models are available on \href{https://huggingface.co/collections/liu-nlp/sift-models}{Hugging Face}.}  All models share the same architecture and parameter count (572M) based on the SmolLM2 architecture \cite{allal2025smollm2smolgoesbig} with the Llama~3.3 tokenizer (128K vocabulary). EN-Base is the English-pretrained model released by \citet{glocker-etal-2025-grow}. FA-Base is trained from scratch on Persian data using the same setup. CPT models are obtained by continued pretraining of EN-Base on different versions of the Persian corpus; all five include replay data drawn from the same English and code corpora used for base model pretraining, comprising approximately 12\% replay data relative to the Persian training tokens, as data replay has been shown to reduce catastrophic forgetting during continued pretraining \citep{scialom-etal-2022-fine}, and preserving source-language factual knowledge is critical to our experimental design.

\subsection{Evaluation} \label{subsec:evaluation}
For each triple \triple{s}{r}{o}, we construct the candidate set by combining the
correct object with the four negative candidates. Each candidate is scored by filling it into each of the five templates for relation $r$ and computing the length-normalized log-probability of the resulting sentence; we average a candidate's scores across the five templates and take the highest-scoring candidate as the model's prediction, so that no individual template determines the outcome.\footnote{We note that results are consistent across individual templates, with per-template accuracy varying by at most 5.2 points for any model. Per-template breakdowns ($\text{mean} \pm \text{std}$) are provided in Appendices~\ref{app:robustness} and~\ref{app:mean-results} (Tables~\ref{tab:main-results-descriptive-hard}--\ref{tab:main-results-random}; Figure~\ref{fig:template-robust}).} We report six metrics. \textbf{Acc\textsubscript{EN}} and \textbf{Acc\textsubscript{FA}} measure top-1 accuracy in English and Persian independently. \textbf{Overall} \cite{goldman2025eclektic} is the fraction of triples answered correctly in both languages simultaneously. \textbf{Transfer} \cite{goldman2025eclektic} is Persian accuracy conditioned on the model answering correctly in English, isolating cases where knowledge acquired in English could have transferred cross-lingually. \textbf{Non-Transfer} is Persian accuracy conditioned on the model answering incorrectly in English, capturing facts the model knows in Persian independently of English. \textbf{RankC} \cite{qi-etal-2023-cross} measures the Spearman rank correlation between the model's candidate orderings in English and Persian. All metrics are computed separately for hard and random candidate sets.

\section{Results and Discussion} \label{sec:results}

Table~\ref{tab:main-results} summarizes fact recall and cross-lingual transfer across all model variants under both hard and random candidate settings. We organize the results around four questions: how much target-language data must be removed to ensure a fact is truly unseen, and whether the removal itself confounds the comparison (\S\ref{subsec:results_removal_enough}); how much factual knowledge transfers from English to Persian under this controlled setting (\S\ref{subsec:limited_transfer}); to what extent does performance reflect factual knowledge rather than shallow associative cues (\S\ref{subsec:shallow-cues}); and how does a fact’s origin, and, related to that, frequency in the English corpus affect its transferability (\S\ref{subsec:freq-transfer}).

Note that the empirical floor of our models is above the 20\% chance level: EN-Base, which has no Persian exposure, reaches 26.0\% in Persian on hard candidates and 31.4\% on random, and FA-Base reaches 32.8\% and 40.0\% in English. We treat these scores as our baseline.

\begin{table*}[h]
\resizebox{\textwidth}{!}{%
\begin{tabular}{l cccccc @{\hspace{1.5em}} cccccc}
\toprule
 & \multicolumn{6}{c}{\textbf{Hard Candidates}} 
 & \multicolumn{6}{c}{\textbf{Random Candidates}} \\
\cmidrule(lr){2-7} \cmidrule(lr){8-13}
\textbf{Model} 
 & Acc\textsubscript{EN}
 & Acc\textsubscript{FA} 
 & Overall
 & Transfer
 & Non-Transfer
 & RankC 
 & Acc\textsubscript{EN}
 & Acc\textsubscript{FA} 
 & Overall
 & Transfer
 & Non-Transfer
 & RankC \\
\midrule
EN-Base & 55.2 & 26.0 & 15.2 & 27.5 & 24.1 &  4.4 
  & 68.8 & 31.4 & 23.2 &   33.7 &   26.3 & 10.7 \\
FA-Base & 32.8 & 35.0 & 12.8 & 39.0 & 33.0 & 11.2 
  & 40.0 & 43.8 & 21.0 &   52.5 &   38.0 & 19.8 \\
CPT-Full & 52.6 & 46.6 & 31.4 & 59.7 & 32.1 & 24.8 
  & 66.8 & 61.6 & 47.2 &   70.7 &   43.4 & 38.8 \\
CPT-CoocRM & 52.8 & 40.0 & 25.8 & 48.9 & 30.1 & 21.8 
  & 66.8 & 54.2 & 41.2 &   61.7 &   39.2 & 34.6 \\
CPT-SubjRM & 51.6 & 32.4 & 20.6 & 39.9 & 24.4 & 14.6 
  & 67.4 & 42.4 & 33.6 &   49.9 &   27.0 & 22.2 \\
CPT-SubjRP & 52.2 & 28.0 & 18.6 & 35.6 & 19.7 & 14.2 
  & 67.6 & 35.2 & 28.0 &   41.4 &   22.2 & 22.5 \\
CPT-RandRM & 54.2 & 47.2 & 33.4 & 61.6 & 30.1 & 25.6 
  & 67.0 & 61.4 & 47.2 &   70.4 &   43.0 & 37.6 \\
\bottomrule
\end{tabular}
}
\caption{Fact recall and cross-lingual transfer across model variants. Hard candidates are semantically close distractors that suppress shallow cues; random candidates are sampled uniformly from the same relation. \textsc{AccEN} and \textsc{AccFA} report top-1 accuracy in English and Persian; \textsc{Overall} is the fraction of triples answered correctly in both. \textsc{Transfer} is Persian accuracy conditioned on English correctness; \textsc{Non-Transfer} conditions on English incorrectness. \textsc{RankC} is the Spearman rank correlation between English and Persian candidate orderings.}
% \caption{Factual recall and cross-lingual transfer across model variants. Metrics are defined in (\S\ref{subsec:evaluation})}
\label{tab:main-results}
\end{table*}

\subsection{How Much Removal Is Enough?}
\label{subsec:results_removal_enough}

To determine the level of target-language data intervention required to ensure a fact is entirely unseen, we evaluate the model's fact retrieval across the removal granularities introduced in \S\ref{subsec:removal}, and we use our random removal control to test whether the
amount of data removed confounds this comparison. Results are in Table~\ref{tab:main-results}. 

\paragraph{Co-occurrence removal is insufficient.}
We find that sentence-level co-occurrence removal is insufficient to guarantee fact absence. Under CPT-CoocRM, which only removes sentences where the subject and object co-occur, the model still predicts many facts correctly in Persian. This could in principle reflect either (1) cross-lingual transfer from English or (2) an insufficient removal strategy that leaves fact signal in the Persian corpus. We argue for the latter. Both Overall and Transfer scores remain substantially higher under CPT-CoocRM than under full subject removal (Overall: 25.8 vs.\ 20.6; Transfer: 48.9 vs.\ 39.9 on hard candidates), despite the same English knowledge being available for transfer in both conditions. More directly, the Non-Transfer score under CPT-CoocRM reaches 30.1\% (hard): the model correctly predicts facts in Persian even when it fails in English,  where cross-lingual transfer is highly unlikely. We hypothesize that the model acquires factual associations from broader document-level context, where the subject and object appear across sentence boundaries.

\paragraph{Full subject removal as the primary intervention.}
Moving to full subject removal, the Non-Transfer score drops from 30.1\% under CPT-CoocRM to 24.4\% under CPT-SubjRM (hard), matching EN-Base's 24.1\%, indicating that removing all sentences containing the subject more effectively mitigates fact exposure from broader contexts. Thus, CPT-SubjRM serves as our primary intervention level, providing a more rigorous setting to keep the fact unseen in the target language without excessive data loss. From a causal perspective, CPT-SubjRM therefore provides our closest approximation to an intervention that removes target-language facts while preserving source-language knowledge.
% While removing entire documents containing both the subject and object would also eliminate this exposure, such an approach discards a massive amount of unrelated, valuable linguistic data.

\paragraph{Adding subjects back in neutral templates degrades performance.} 
As CPT-SubjRM removes the Persian surface form of the subject entity completely, we evaluate CPT-SubjRP to observe the empirical effect of preserving surface-form exposure via neutral templates. The results show that this approach actually degrades performance, achieving only 28.0\% (hard) and 35.2\% (random) compared to CPT-SubjRM's 32.4\% and 42.4\%. The near identical rank correlation scores (14.2--14.6 on hard, 22.2--22.5 on random) indicate that the degradation is unlikely to stem from a difference in cross-lingual alignment. We propose two explanations for this degradation: first, repeatedly seeing an entity in neutral Persian sentences may cause the model to build a shallow, Persian-only representation of that entity, weakening the factual knowledge it had already acquired from English. It could also be an effect of duplicate %or near-duplicate
data, which has been shown to degrade model quality
\citep{lee-etal-2022-deduplicating} and may lead to reproducing memorized templates rather than retrieving facts \citep{carlini2023quantifying}. 

 % Two explanations likely account for this degradation: first, repeatedly seeing an entity in neutral Persian sentences may cause the model to build a shallow, Persian-only representation of that entity, weakening the factual knowledge it had already acquired from English; second, replacing removed sentences with a fixed pool of generic templates introduces unnatural repetition into the training data, which inherently degrades the quality of the learned representations.

\paragraph{Removal volume alone does not explain the drop.}
CPT-RandRM removes 7.79\% of Persian tokens, comparable to the 8.50\% removed under
CPT-SubjRM, while leaving all factual evidence in place. It reaches 47.2\% Persian accuracy on hard candidates, statistically indistinguishable from CPT-Full's 46.6\%, whereas CPT-SubjRM falls to 32.4\% under the same evaluation. Removing a similar amount of text therefore leaves accuracy almost unchanged, as long as the factual evidence stays in. We therefore attribute the drop under CPT-SubjRM to the removed factual evidence rather than to the reduced corpus size.

\subsection{Limited Cross-Lingual Transfer}
\label{subsec:limited_transfer}
% English accuracy remains stable across all CPT models (51-53\% hard, 66-68\% random), confirming that the 12\% English replay successfully preserves source-language knowledge.

Having established CPT-SubjRM as our primary intervention, in which no Persian-language exposure to the entities remains, we now use it to quantify the degree of cross-lingual fact transfer. Following the evaluation framework of ECLeKTic \citep{goldman2025eclektic}, we ask: to what extent can the model retrieve English-acquired facts through Persian?

\paragraph{There is little fact transfer.} 
The Persian hard-candidate accuracy under CPT-SubjRM approximates the probability of correct Persian prediction when Persian factual evidence and shallow heuristic cues are minimized. Under the causal interpretation discussed in Appendix~\ref{app:causal-effect-estimation}, performance above the empirical floor in this setting therefore provides evidence of cross-lingual fact transfer from English. As shown in Table~\ref{tab:main-results}, CPT-SubjRM on hard candidates achieves an Overall score of only 20.6\%, indicating that the model rarely succeeds in both languages simultaneously. Its Transfer score reaches 39.9\%, suggesting that some English-acquired knowledge is accessible in Persian, although most is not. This is consistent with \citet{goldman2025eclektic}, who find that even large-scale models hardly share knowledge across languages. A per-relation breakdown reveals substantial variation in both fact recall and the effect of our intervention across relation types (Appendix~\ref{app:per-relation-recall}).

\paragraph{Distinguishing transfer from difficulty confounds.} 
To further assess whether this gap reflects genuine transfer rather than a difficulty confound, we compare the Transfer and Non-Transfer accuracies. CPT-SubjRM shows a difference of 15.5 percentage points between the two (39.9\% vs.\ 24.4\%), while EN-Base shows virtually no such gap (27.5\% vs.\ 24.1\%). If the gap were driven solely by certain facts being inherently easier, it would appear in both models; its absence in EN-Base suggests that the gap in CPT-SubjRM reflects a limited degree of cross-lingual transfer enabled by continued pretraining. Nonetheless, Transfer and Non-Transfer accuracies both remain low, and the Non-Transfer accuracy (24.4\%) falls below the 26.0\% empirical floor, indicating that the model's ability to transfer knowledge into Persian is, at most, limited. 
% , while present, is far from reliable.

\paragraph{Rank-based evaluation.}
Beyond binary correctness, we use RankC to examine whether cross-lingual transfer is also reflected in the model's full ranking over candidates. EN-Base, which has no Persian exposure, serves as the baseline at 4.4 on hard candidates. CPT-SubjRM obtains a RankC of 14.6 on hard candidates. Although both values remain low relative to the theoretical maximum (100), the increase suggests that continued pretraining on Persian text partially aligns candidate preferences across languages, even for entities absent from the Persian training data.

\subsection{Precise Knowledge or Shallow Heuristics?}\label{subsec:shallow-cues}
Comparing hard and random candidate settings reveals the extent to which models rely on shallow associative cues. Across all models, the random-candidate accuracy substantially exceeds the hard-candidate accuracy in both languages. For CPT-SubjRM, the Persian accuracy gap is 10 percentage points (42.4\% vs.\ 32.4\%), while the English accuracy gap is 15.8 points (67.4\% vs.\ 51.6\%), indicating that surface-level cues, such as associating a subject from a particular cultural region with geographically related objects, inflate performance in both languages, not only in the target language. Crucially, this effect persists even when conditioning on English knowledge: the Transfer score for CPT-SubjRM drops from 49.9\% with random candidates to 39.9\% with hard candidates, and for CPT-Full from 70.7\% to 59.7\%. This suggests that a portion of the apparent cross-lingual transfer measured under easier conditions is itself attributable to shallow heuristics rather than precise fact retrieval \citep{saynova-etal-2025-fact}. The hard candidate setting, which includes distractor candidates that are semantically and culturally close to the correct answer, reduces the contribution of such heuristics to the measured transfer scores.

\subsection{Entity Frequency and Provenance Affect Transferability}
\label{subsec:freq-transfer}

\begin{figure}[t]
\centering
\includegraphics[width=\columnwidth]{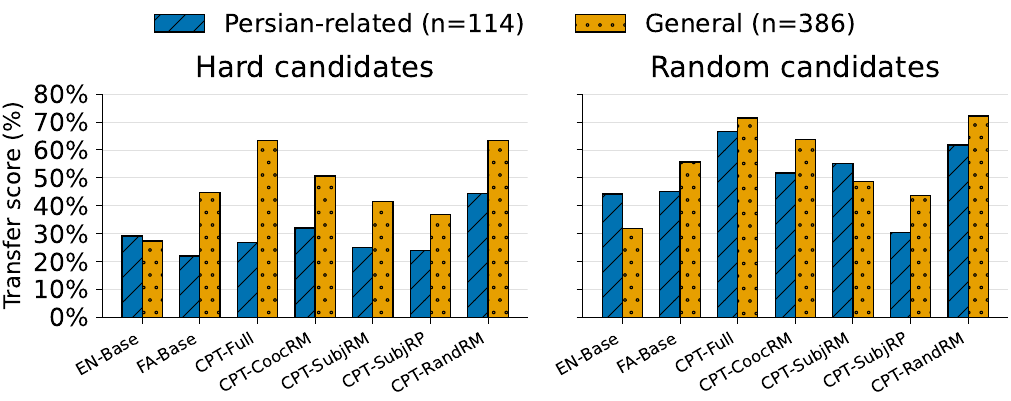}
\caption{Transfer score by provenance category under hard and random candidate settings. Counts (n) refer to the number of triples per category (500 total).}
\label{fig:transfer-cultural}
\end{figure}

\begin{figure*}[h]
    \centering
        \includegraphics[width=1\linewidth]{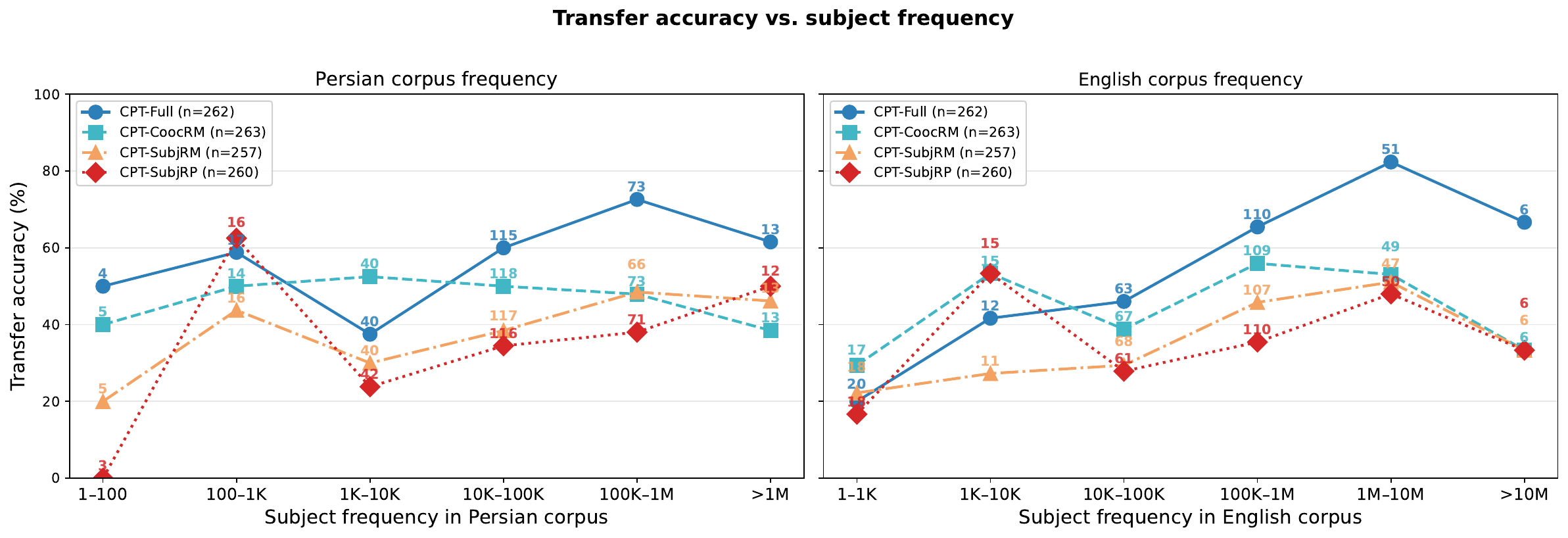}
    \caption{Transfer accuracy (\%) on hard candidates as a function of subject frequency in the Persian (left) and English (right) pretraining corpora. Numbers above points indicate subjects per bin.}
    \label{fig:freq-transfer}
\end{figure*}

Fact recall scales with entity frequency in monolingual settings \citep{kandpal2023large}, and a similar relationship has been observed for cross{\FOO}lingual transfer \citep{liu-etal-2025-tracing, zhang-etal-2025-cross}, but without isolating source{\FOO}language frequency from direct target-language learning. Because our training data intervention removes all subject mentions from the Persian corpus, we can isolate the role of English frequency more cleanly. 

\begin{table}[t]
\centering
\setlength{\tabcolsep}{4pt}
\resizebox{0.5\textwidth}{!}{%
\begin{tabular}{l r r c c}
\toprule
 & \multicolumn{2}{c}{\textbf{Median freq.}} 
 & \multicolumn{2}{c}{\textbf{\% subjects in EN}} \\
\cmidrule(lr){2-3} \cmidrule(lr){4-5}
 & EN & FA & {$<$100} & {$<$1K} \\
\midrule
General Facts\   ($n$=324) & 196K & 35K & 0 & 1 \\
Persian{\FOO}related Facts\ ($n$=113) & 284  & 40K & 36 & 65 \\
\bottomrule
\end{tabular}
}
\caption{Median subject frequency in the English and Persian pretraining corpora by provenance category. Counts (n) refer to unique subjects (437 total), which are fewer than the 500 triples reported in (\S\ref{subsec:factual_triples}) since a subject may appear in multiple triples across relations.}
\label{tab:frequency-by-culture}
\end{table}

% \citet{kandpal2023large} established that factual recall scales with entity frequency in monolingual settings. We examine whether a similar frequency dependence governs cross-lingual transfer. 

\paragraph{English frequency gates transfer.}
Figure~\ref{fig:freq-transfer} shows Transfer accuracy as a function of subject frequency in both the English and Persian pretraining corpora. For CPT-SubjRM, Transfer accuracy increases steadily with English corpus frequency up to the 1M–10M bin: entities appearing fewer than 1{,}000 times in English are transferred near the empirical floor, while entities in the 100K--10M range reach 46--51\%. This confirms that the strength of the English-side representation is a bottleneck for cross-lingual transfer, entities that the model encountered more frequently in English are more likely to be retrievable in Persian after continued pretraining, even when those entities were never seen in Persian. However, even at the highest English frequencies, CPT-SubjRM remains well below CPT-Full (51\% vs.\ 82\% in the 1M--10M bin), indicating that English exposure is insufficient for reliable transfer and that direct Persian exposure contributes substantially.
 
\paragraph{Persian frequency as an indirect signal.}
On the Persian frequency side, CPT-SubjRM still shows a positive trend despite having all subject mentions removed from the Persian corpus. Since Persian corpus frequency reflects the general prominence of an entity, this correlation likely arises indirectly: globally prominent entities tend to be well-represented in English as well, providing a stronger source{\FOO}language representation to transfer from. The gap between CPT-Full and CPT-SubjRM is substantial across the frequency range and largest in the well-populated high-frequency bins (e.g., 100K--1M: 73\% vs.\ 49\%), showing that while cross-lingual transfer accounts for the majority of performance, direct learning from Persian contributes a further 24-percentage-point gain.
%at these frequency levels.

\paragraph{Impact of the origin of facts. } 
The frequency bottleneck manifests also across provenance categories. As Table~\ref{tab:frequency-by-culture} shows, Persian{\FOO}related subjects (e.g., Persian historical figures) have a median English frequency of only 284, roughly 690 times lower than general subjects (median 196K), with 65\% appearing fewer than 1{,}000 times and 36\% fewer than 100 times. Persian-side frequencies, by contrast, are comparable across categories (40K vs.\ 35K median). The consequence is visible in Figure~\ref{fig:transfer-cultural}: on hard candidates, CPT-SubjRM transfers general facts at 41.5\% but Persian{\FOO}related facts at only 25.0\%. EN-Base shows no such gap (27.4\% vs.\ 29.2\%), confirming that the disparity is not due to Persian{\FOO}related facts being inherently harder but rather to their sparse English representation. The low English accuracy of CPT-SubjRM on Persian{\FOO}related facts (21.1\% vs.\ 60.6\% on general) corroborates this: the model simply lacks the source-language knowledge from which transfer could occur. On random candidates the ordering reverses (48.7\% general vs. 55.2\% Persian-related), as Persian names offer strong associative cues that randomly sampled distractors rarely defeat. EN-Base, with no Persian exposure, shows the same reversal (31.8\% vs. 44.2\%), but the hard-candidate setting strips away these heuristics and exposes the underlying frequency-driven limitation.

\section{Conclusion and Future Work}
We presented an intervention-based framework for studying cross-lingual factual knowledge transfer during continued pretraining. By systematically removing specific facts from Persian training data and probing the resulting model, we established causal evidence that cross-lingual transfer from English exists but is limited. When our interventions remove the facts from the Persian data and remove access to shallow heuristics that make predictions easier, the model transfers only a small fraction of its English-acquired knowledge into Persian, with transfer largely confined to general facts whose subjects are frequent in the English corpus: entities with low English exposure transfer at near-chance levels, while even highly frequent entities have limited performance, indicating that English-side representation strength is necessary but insufficient for reliable transfer.

One factor potentially limiting transfer between English and Persian is the difference in script. Unlike language pairs that share the Latin alphabet, Persian's Arabic-derived script offers no orthographic overlap with English, and thereby no token overlap, which may hinder the model's ability to align entity representations across languages. A promising direction for future work is to incorporate transliterated data into the pretraining corpus, as proposed by \citet{liu-etal-2024-translico, liu-etal-2025-transliterations}, to test whether bridging the script gap enables stronger cross-lingual fact transfer even in our intervention-based setup.

\section*{Limitations}
Our study is limited to a single target language, Persian, which uses a non-Latin script with no orthographic overlap with English. Extending this framework to languages that share the Latin script with English would help determine whether script similarity facilitates stronger cross-lingual fact transfer. Similarly, including additional non-Latin-script languages would test the generality of our findings beyond the English--Persian pair. However, both extensions are challenging: our approach requires natively written templates, manually curated triples with country-specific facts, and hard candidate sets designed by speakers with expertise in the target language, making each new language a substantial annotation effort rather than a simple translation task. 

In addition, our models are limited to 572M parameters. \citet{goldman2025eclektic} find that scaling model size improves overall fact recall but does not substantially improve cross-lingual transfer, but their conclusion is based on observational comparisons without training data interventions. Whether the same pattern holds under controlled fact removal remains an open question that would require training larger models. %, which was beyond our computational budget.

\section*{Acknowledgments}

We thank the anonymous reviewers for their constructive feedback and insightful suggestions. 
This research was supported by TrustLLM funded by Horizon Europe GA 101135671 and the Graduate School in Computer Science (CUGS). 
It was partially supported by the Wallenberg AI, Autonomous Systems and Software Program (WASP) funded by the Knut and Alice Wallenberg Foundation. 
The computations were enabled by the Berzelius resource provided by the Knut and Alice Wallenberg Foundation at the National Supercomputer Centre.

\bibliography{custom}

\appendix

\input{appendix_causality}

\section{Triple Creation Criteria} \label{app:triples-criteria}

\paragraph{Subject diversity} Language models tend to produce the same prominent entities across multiple relations, for instance, generating ``France'' under \semrel{capital}, \semrel{continent}, and \semrel{official\_language}, or ``Albert Einstein'' under \semrel{field\_of\_work}, \semrel{place\_of\_birth}, and \semrel{country\_of\_citizenship}. Because our data intervention removes sentences based on subject-object co-occurrence, a subject appearing in many relations would trigger broad removal of text that is unrelated to the specific fact being tested. We therefore curated the final selection to maximize subject diversity across relations.

\paragraph{Temporal diversity} The generated candidates were heavily skewed toward historical figures. We deliberately included contemporary public figures to ensure the dataset is not biased toward entities that are disproportionately represented in encyclopedic sources.

% Figure belongs to "factual recall" section below; moved here so it floats to somewhere close to it
\begin{figure*}[h]
    \centering
    \includegraphics[width=1\linewidth]{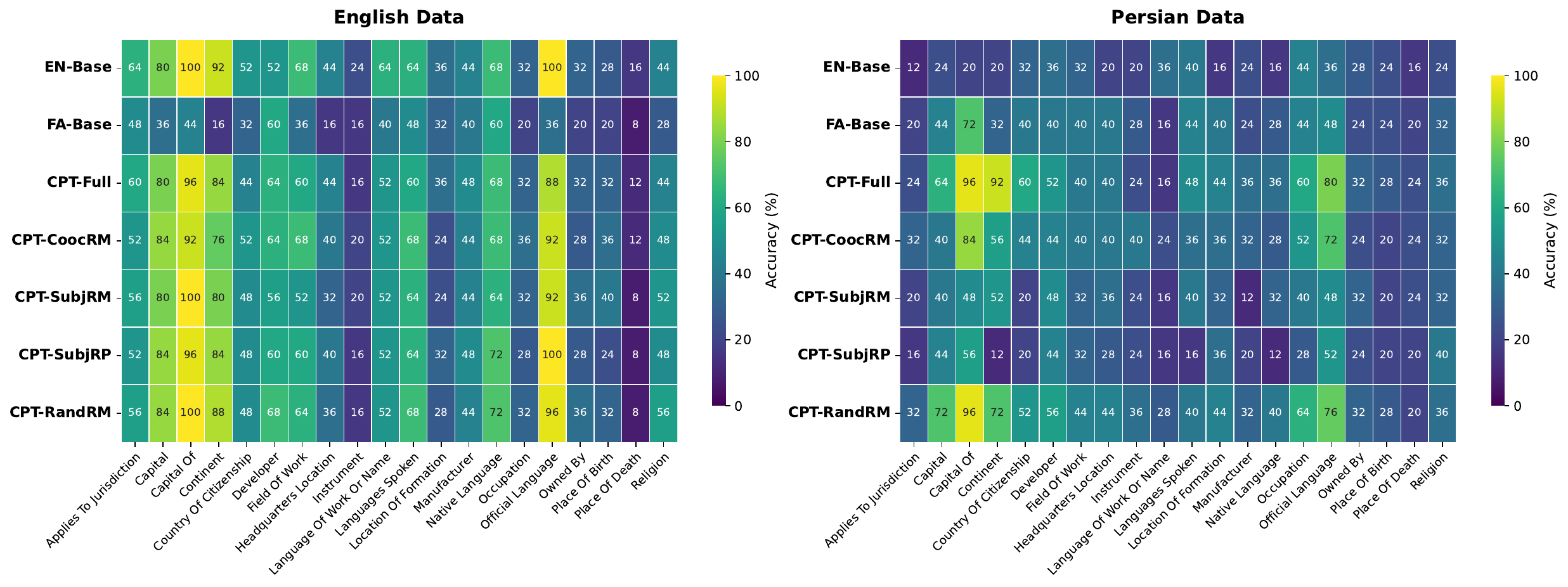}
    \caption{Per-relation accuracy (\%) on hard candidates for English (left) and Persian (right) evaluation.}
    \label{fig:heatmap}
\end{figure*}

\paragraph{Distinct subject-object surface forms} Some generated triples had near-identical subject and object strings e.g., \triple{Germany}{official\_language}{German}, which causes unintended removals during filtering and makes evaluation trivially solvable through string overlap rather than factual knowledge. We replaced such triples with pairs whose surface forms are clearly distinct e.g., \triple{Austria}{official\_language}{German}.

\paragraph{Person name variants} For triples with person subjects, we identified all name variants likely to appear in natural text to ensure comprehensive matching during the data intervention. Full names typically appear at the start of a document, while shortened forms are used thereafter. We handled this by generating variant lists, for example, ``Pablo Picasso'' yields the additional variant ``Picasso''. These variants were verified against both English and Persian Wikipedia pages.

\paragraph{Exhaustive object enumeration} Some subjects have more than one correct object for a given relation, for example, Stephen Hawking's \semrel{fields\_of\_work} include both ``cosmology'' and ``physics''. In such cases, we list all valid objects in the dataset. This is necessary because our filtering pipeline only removes facts that are explicitly listed: if we listed ``cosmology'' but not ``physics'', sentences linking ``Hawking'' to ``physics'' would remain in the training corpus, and the model could appear to transfer knowledge that it actually learned directly from the unfiltered Persian data. Similarly, when an object can be written in multiple ways in Persian, we include all surface form variants to ensure complete removal.

\paragraph{Object diversity within relations} For each relation, we selected objects that cover a broad range of values rather than concentrating on the most frequent ones. For instance, the \semrel{religion} relation includes Christianity, Islam, Zoroastrianism, Mazdakism, Judaism, Hinduism, Buddhism, and Sikhism; the \semrel{capital} relation includes both well-known capitals (e.g., Paris, Tokyo) and less prominent ones.

\paragraph{Corpus frequency control} Some triples occurred so frequently in the Persian FineWeb-2 corpus that removing them would have distorted the language distribution of the training data; we replaced these with less frequent alternatives. For example, subjects such as Iran, China, Russia, Samsung, and Microsoft alone accounted for approximately 6\% of all tokens under full subject removal.

\section{Training Data Replacement Templates}\label{app:replace}
In our template replacement condition (\S\ref{subsec:removal}), instead of removing sentences containing a subject entity, we replace them with neutral substitute sentences that preserve the subject's surface form without exposing the model to the target factual associations. We authored a pool of 100 neutral Persian sentences per entity type (e.g., \semrel{person}, \semrel{country}, \semrel{city}), each containing a ``{\{subject\}}'' placeholder, shared across relations with the same subject type. Each template satisfies three constraints: (1) it must be true for every subject in the entity-type group; (2) it must not contain vocabulary that could indirectly encode the target relation, enforced via per-type forbidden-vocabulary lists; and (3) templates vary in length, position, and style to avoid repetitive contexts. During filtering, each targeted sentence is replaced by a randomly drawn template with the placeholder filled by the entity's Persian surface form.

\section{How Does Fact Recall Vary By Relation?}\label{app:per-relation-recall}

Figure~\ref{fig:heatmap} reveals substantial variation in fact recall across relations. Relations expressed through frequent and formulaic patterns in the training data, such as \semrel{capital\_of}, \semrel{continent}, and \semrel{official\_language}, are the easiest across all models, with EN-Base reaching 92--100\% in English. For \semrel{continent}, the constrained answer space (six possible objects) further contributes to high accuracy. These relations also show the largest drop under our intervention: Persian accuracy falls 32--48 points from CPT-Full to CPT-SubjRM (e.g., \semrel{capital\_of}: 96\%$\to$48\%, \semrel{continent}: 92\%$\to$52\%), confirming that CPT-Full's performance on these relations was largely attributable to direct learning from the Persian corpus.

Conversely, person-attribute relations such as \semrel{occupation}, \semrel{developer}, and \semrel{languages\_spoken} show smaller drops under subject removal (4--20 points) and remain above the empirical floor under CPT-SubjRM (40--48\%). However, for relations with few possible answers, such as \semrel{religion} (8 unique objects), this may partly reflect the constrained answer space rather than robust recall.

\begin{figure*}[h]
\centering
\includegraphics[width=\textwidth]{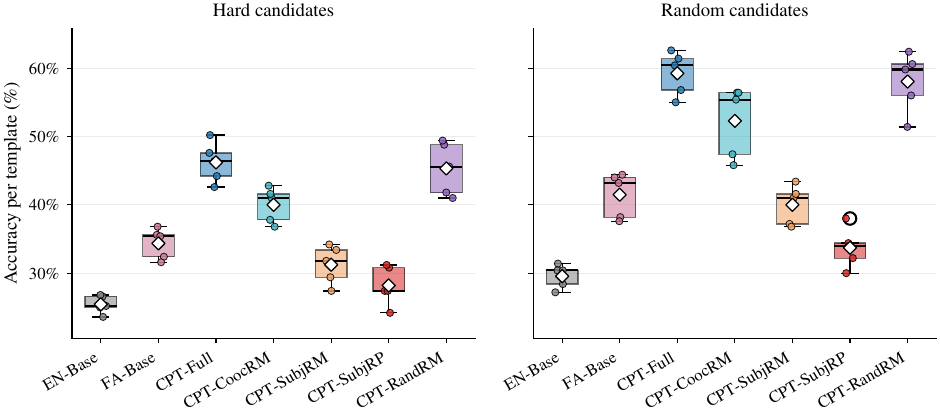}
\caption{Per-template accuracy distribution for each model 
under hard (left) and random (right) candidate settings. 
Each box summarizes accuracy across 5 evaluation templates; 
individual template scores are shown as points, and diamonds 
indicate the mean. The narrow spread (std $\leq$ 5.2pp) 
confirms that results are robust to template phrasing.}
\label{fig:template-robust}
\end{figure*}

At the other end, \semrel{place\_of\_death} remains near or below the empirical floor in both languages across all models, with no drop from CPT-Full to CPT-SubjRM. This fact type was not acquired even from the full Persian data. Notably, for \semrel{language\_of\_work\_or\_name}, continued pretraining actually degrades Persian accuracy from 36\% (EN-Base) to 16\% (CPT-SubjRM).

\section{Robustness to Different Templates}\label{app:robustness}

Figure~\ref{fig:template-robust} shows the distribution of per-template accuracy across all five evaluation templates for each model under both hard and random candidate settings. The narrow spread of scores confirms that our results are robust to variation in template phrasing: the standard deviation across templates is at most 5.2 percentage points for any model, and the relative ordering of models is consistent across all templates. This indicates that the patterns reported in the main results (\S\ref{sec:results}) are not driven by idiosyncratic properties of individual templates.

\section{Prompt for Hard Distractor Candidate Generation}
\label{app:candidate-prompt}

The prompt in Figure~\ref{fig:candidate-prompt} was used to generate hard distractor candidates for each relation type. It instructs the model to produce four incorrect but plausible candidates per triple, following relation-specific criteria designed to maximize semantic and cultural proximity to the correct answer. For example, candidates for geographic relations are drawn from neighboring or commonly confused locations, while candidates for person-attribute relations exclude all known correct values and favor entities from the same cultural or professional domain. All generated candidates were subsequently verified by a native Persian speaker to ensure factual incorrectness and sufficient difficulty.

\section{Mean Results on Different Templates} \label{app:mean-results}

Tables~\ref{tab:main-results-descriptive-hard}--\ref{tab:main-results-random} report mean accuracy and standard deviation across the five evaluation templates for all models, broken down by template format (descriptive and question answering) and candidate setting (hard and random). The results are broadly consistent across both template formats, with the same relative ordering of models and the same overall patterns observed in the main results. One notable difference is that CPT-SubjRP slightly outperforms CPT-SubjRM under the question answering format (Tables~\ref{tab:main-results-hard} and~\ref{tab:main-results-random}), by 2.9 points on hard and 5.7 points on random candidates, reversing the pattern seen with descriptive templates. The reversal exceeds template-level variation, indicating that the relative ordering of these two variants is sensitive to template format. Overall, the consistency across template formats confirms that our findings are not sensitive to the specific evaluation format used.

\onecolumn
\clearpage
\refstepcounter{figure} % Without this, the label will be that of the appendix
\addtocounter{figure}{-1} % Without this, the label will be off by 1
\label{fig:candidate-prompt}
\begin{tcolorbox}[colback=gray!3, colframe=gray!60, breakable, fontupper=\small\ttfamily]
You are generating object\_candidates for knowledge base triples. Each triple has a subject, relation, and object(s). Generate 4 wrong candidates that do NOT include any correct answer. Follow these rules per relation:

\medskip
\textbf{1. capital:} Use the largest non-capital cities of the same country, capitals of neighboring countries, cities with similar spelling to the correct capital, and cities commonly mistaken as the capital. For historical empires, use major cities within the empire's territory that were never its capital and cities that served as capital of a predecessor or successor state.

\textbf{2. capital\_of:} Use geographically neighboring countries sharing a border, countries sharing the same language or colonial history, countries commonly confused with each other, countries the city has historical ties to, and countries on the same continent with similar characteristics.

\textbf{3. continent:} Use the geographically closest continents, continents the country is commonly misassigned to, and for transcontinental countries the other continent they straddle. Replace distant implausible continents with more confusing ones.

\textbf{4. country\_of\_citizenship:} Exclude all countries the person ever held citizenship in. Use countries the person lived in or worked in but never held citizenship of, neighboring or culturally linked countries, countries sharing the same language, countries the person is commonly but incorrectly associated with, and countries where the person had significant career activity. If there is any doubt about citizenship, exclude the candidate.

\textbf{5. developer:} Use direct competitors making the same category of product, companies that make the closest rival product, and parent or subsidiary companies in the same tech ecosystem. For Chinese apps use other major Chinese tech companies, and for Iranian apps use other Iranian tech ecosystem companies. Never include the correct developer under any alternate name.

\textbf{6. field\_of\_work:} Exclude any field the person genuinely worked in even partially, and exclude overly general fields that could apply to anyone. Use fields adjacent to but distinct from the correct one. For highly multi-disciplinary figures, use clearly anachronistic or unrelated fields. Easy candidates are acceptable, and if there is any doubt whether a field applies, exclude it.

\textbf{7. headquarters\_location:} Use other cities in the same metro area, cities where the company has major offices or factories, and the founder's city if different from the headquarters. For Japanese companies use other major Japanese industrial cities, and for Iranian institutions use nearby cities or functionally similar ones.

\textbf{8. instrument:} Exclude every instrument the musician is known to play. For Iranian multi-instrumentalists use only Western instruments, and for Iranian musicians who only play Western instruments, Iranian instruments are valid wrong candidates. Use instruments commonly associated with the same genre and instruments played by the musician's bandmates or frequent collaborators.

\textbf{9. languages\_spoken:} Exclude every language the person speaks, not just the listed answer. Use languages of countries the person lived in but confirmed they do not speak, closely related languages to their known ones, languages of the person's geographic or cultural region, and languages of countries where the person had significant career activity. If there is any doubt whether the person speaks a candidate language, exclude it.

\textbf{10. native\_language:} Exclude every language the person is native in or spoke from childhood. Use languages of the country the person is most associated with, closely related languages, and languages the person learned later in life but is not native in. If there is any doubt about native status, exclude it.

\textbf{11. official\_language:} Exclude all co-official languages of the country. Use the largest minority languages spoken within the country, languages of neighboring countries, former colonial or administrative languages if not currently official, languages historically associated with the country, and languages commonly but incorrectly assumed to be official.

\textbf{12. occupation:} Exclude all real occupations including secondary ones and exclude occupations too general that could plausibly apply. For athletes use other sports or athletic roles, and for performers use nearby performance roles. Use occupations of close collaborators or contemporaries and occupations the person is commonly but incorrectly described as.

\textbf{13. religion:} Use religions historically present in the person's region of origin, religions the person studied or was influenced by but did not follow, the dominant religion of the country the person is associated with, and religions commonly confused with the correct one in the region. Do not include denominational distinctions.

\textbf{14. applies\_to\_jurisdiction:} Use jurisdictions with genuine historical or geographic connections to the law, countries that passed similar laws or policies, the other party in a bilateral treaty if it is not also a correct answer, and predecessor or successor states of the correct jurisdiction. Do not include member states of the actual jurisdiction.

\textbf{15. manufacturer:} Use direct competitors making the same or very similar product, companies making the closest rival product in the same segment, and companies commonly confused as the manufacturer. For Iranian products use other Iranian manufacturers in the same sector. Never use a company from an unrelated industry.

\textbf{16. owned\_by:} Use companies that attempted or were rumored to acquire the subject, companies that own the closest competing product, the previous owner before an acquisition, and companies in the same investment portfolio or sector. For Iranian entities use same-sector Iranian organizations. Never use a company from an unrelated sector.

\textbf{17. language\_of\_work\_or\_name:} Use languages historically tied to the work's setting or author's world, the language of the country where the work is primarily set, and languages the author spoke but did not write in. For works with significant passages in another language, exclude that language. Use the language of the most famous translation and languages commonly but incorrectly assumed for the work.

\textbf{18. place\_of\_birth:} Use other cities in the same country associated with the person, cities the person grew up in or is most famous for, and cities commonly but incorrectly believed to be the birthplace. For Iranian subjects use other major cities in the same province or region. Use birthplaces of closely related or commonly confused figures and the city the person is most associated with if different from the birthplace.

\textbf{19. place\_of\_death:} Use cities the person lived in near the end of their life, cities of exile or emigration, and the person's birthplace if commonly confused with place of death. For Iranian figures who died abroad include Tehran and the exile city. For historical scholars use major cities of their intellectual world, and use cities where the person had their last major public appearance.

\textbf{20. location\_of\_formation:} Use the organization's current headquarters if different from the formation location, cities where key negotiations or predecessor events took place, and cities of predecessor organizations. For Iranian institutions use cities with similar functional profiles. Use cities commonly but incorrectly believed to be the founding location and cities where the organization held its first major event.

\medskip
\textbf{General rules:} Never include any correct answer in candidates. If a triple has multiple correct answers, exclude all of them. Candidates must be of the same type as the correct answer.
\end{tcolorbox}
\captionof{figure}{The full prompt used to generate hard distractor candidates for each relation type}
\vspace{1cm}

\begin{table*}[h]
\centering
\begin{tabular}{l cccccc}
\toprule
\textbf{Model} 
 & Acc\textsubscript{EN}
 & Acc\textsubscript{FA} 
 & Overall
 & Transfer
 & Non-Transfer
 & RankC \\
\midrule
EN-Base     & 51.1$\pm$4.4 & 25.4$\pm$1.3 & 13.3$\pm$1.7 & 25.9$\pm$1.3 & 24.8$\pm$3.6 &  4.3$\pm$1.3 \\
FA-Base     & 31.8$\pm$0.7 & 34.4$\pm$2.2 & 12.5$\pm$1.1 & 39.3$\pm$3.6 & 32.0$\pm$2.1 & 11.7$\pm$1.8 \\
CPT-Full    & 49.5$\pm$3.3 & 46.2$\pm$3.0 & 28.2$\pm$2.7 & 57.0$\pm$3.4 & 35.6$\pm$2.5 & 21.2$\pm$2.1 \\
CPT-CoocRM  & 50.6$\pm$3.6 & 40.0$\pm$2.6 & 24.3$\pm$2.4 & 48.1$\pm$3.9 & 31.7$\pm$2.3 & 19.8$\pm$1.8 \\
CPT-SubjRM  & 49.5$\pm$3.5 & 31.2$\pm$2.8 & 18.7$\pm$2.3 & 37.8$\pm$3.5 & 24.9$\pm$2.0 & 12.7$\pm$3.0 \\
CPT-SubjRP  & 49.9$\pm$3.9 & 28.2$\pm$2.9 & 16.1$\pm$1.5 & 32.4$\pm$2.8 & 24.0$\pm$3.6 & 10.4$\pm$3.8 \\
CPT-RandRM  & 51.2$\pm$3.0 & 45.3$\pm$3.9 & 29.4$\pm$2.7 & 57.5$\pm$3.6 & 32.4$\pm$5.0 & 22.8$\pm$1.5 \\
\bottomrule
\end{tabular}
\caption{Fact recall and cross-lingual transfer across model variants using a descriptive template with hard candidates. Hard candidates are semantically close distractors that suppress shallow cues. \textsc{AccEN} and \textsc{AccFA} report top-1 accuracy in English and Persian; \textsc{Overall} is the fraction of triples answered correctly in both. \textsc{Transfer} is Persian accuracy conditioned on English correctness; \textsc{Non-Transfer} conditions on English incorrectness. \textsc{RankC} is the Spearman rank correlation between English and Persian candidate orderings.}
\label{tab:main-results-descriptive-hard}
\end{table*}

\begin{table*}[h]
\centering
\begin{tabular}{l cccccc}
\toprule
\textbf{Model} 
 & Acc\textsubscript{EN}
 & Acc\textsubscript{FA} 
 & Overall
 & Transfer
 & Non-Transfer
 & RankC \\
\midrule
EN-Base     & 51.2$\pm$0.7 & 24.3$\pm$0.8 & 14.8$\pm$1.1 & 29.0$\pm$1.9 & 19.3$\pm$0.9 &  3.1$\pm$1.3 \\
FA-Base     & 29.6$\pm$0.9 & 30.5$\pm$0.9 &  8.7$\pm$0.2 & 29.5$\pm$0.9 & 31.0$\pm$1.1 &  6.6$\pm$0.6 \\
CPT-Full    & 50.6$\pm$1.5 & 36.4$\pm$1.1 & 23.6$\pm$0.8 & 46.6$\pm$1.6 & 25.9$\pm$1.8 & 23.3$\pm$1.5 \\
CPT-CoocRM  & 51.4$\pm$0.5 & 32.0$\pm$1.3 & 21.1$\pm$1.1 & 41.0$\pm$1.8 & 22.3$\pm$1.0 & 22.4$\pm$2.0 \\
CPT-SubjRM  & 50.4$\pm$1.2 & 28.2$\pm$0.8 & 18.0$\pm$0.7 & 35.7$\pm$0.6 & 20.5$\pm$0.9 & 14.5$\pm$1.8 \\
CPT-SubjRP  & 49.6$\pm$1.8 & 31.1$\pm$0.6 & 19.6$\pm$1.3 & 39.5$\pm$1.6 & 22.8$\pm$1.2 & 16.4$\pm$2.0 \\
CPT-RandRM  & 52.1$\pm$0.8 & 36.0$\pm$0.7 & 23.9$\pm$0.7 & 45.8$\pm$1.4 & 25.3$\pm$0.9 & 21.2$\pm$1.9 \\
\bottomrule
\end{tabular}
\caption{Fact recall and cross-lingual transfer across model variants using a question answering template with hard candidates. Hard candidates are semantically close distractors that suppress shallow cues. \textsc{AccEN} and \textsc{AccFA} report top-1 accuracy in English and Persian; \textsc{Overall} is the fraction of triples answered correctly in both. \textsc{Transfer} is Persian accuracy conditioned on English correctness; \textsc{Non-Transfer} conditions on English incorrectness. \textsc{RankC} is the Spearman rank correlation between English and Persian candidate orderings.}
\label{tab:main-results-hard}
\end{table*}

\begin{table*}[h]
\centering
\begin{tabular}{l cccccc}
\toprule
\textbf{Model} 
 & Acc\textsubscript{EN}
 & Acc\textsubscript{FA} 
 & Overall
 & Transfer
 & Non-Transfer
 & RankC \\
\midrule
EN-Base     & 66.2$\pm$2.2 & 29.6$\pm$1.7 & 20.8$\pm$1.8 & 31.4$\pm$1.9 & 25.9$\pm$4.7 & 10.0$\pm$2.2 \\
FA-Base     & 37.6$\pm$1.6 & 41.5$\pm$3.3 & 19.0$\pm$2.3 & 50.5$\pm$4.8 & 36.0$\pm$3.1 & 16.7$\pm$1.2 \\
CPT-Full    & 65.2$\pm$3.4 & 59.2$\pm$3.2 & 44.6$\pm$4.3 & 68.4$\pm$5.2 & 42.1$\pm$2.5 & 35.9$\pm$2.0 \\
CPT-CoocRM  & 65.2$\pm$2.7 & 52.3$\pm$5.2 & 39.7$\pm$5.0 & 60.9$\pm$7.0 & 36.0$\pm$2.4 & 31.3$\pm$2.8 \\
CPT-SubjRM  & 65.5$\pm$2.6 & 40.0$\pm$2.9 & 30.9$\pm$2.9 & 47.2$\pm$3.9 & 26.2$\pm$2.6 & 20.2$\pm$1.9 \\
CPT-SubjRP  & 63.9$\pm$3.7 & 33.7$\pm$3.0 & 24.3$\pm$1.7 & 38.2$\pm$4.2 & 26.0$\pm$2.0 & 18.8$\pm$3.2 \\
CPT-RandRM  & 65.5$\pm$2.4 & 58.0$\pm$4.4 & 44.2$\pm$4.8 & 67.4$\pm$5.6 & 40.0$\pm$1.9 & 34.3$\pm$2.7 \\
\bottomrule
\end{tabular}
\caption{Fact recall and cross-lingual transfer across model variants using a descriptive template with random candidates. Random candidates are sampled uniformly from the same relation. \textsc{AccEN} and \textsc{AccFA} report top-1 accuracy in English and Persian; \textsc{Overall} is the fraction of triples answered correctly in both. \textsc{Transfer} is Persian accuracy conditioned on English correctness; \textsc{Non-Transfer} conditions on English incorrectness. \textsc{RankC} is the Spearman rank correlation between English and Persian candidate orderings.}
\label{tab:main-results-descriptive-random}
\end{table*}

\begin{table*}[h]
\centering
\begin{tabular}{l cccccc}
\toprule
\textbf{Model} 
 & Acc\textsubscript{EN}
 & Acc\textsubscript{FA} 
 & Overall
 & Transfer
 & Non-Transfer
 & RankC \\
\midrule
EN-Base     & 65.6$\pm$1.3 & 26.4$\pm$0.7 & 18.5$\pm$0.7 & 28.2$\pm$1.0 & 23.2$\pm$2.6 &  9.4$\pm$1.1 \\
FA-Base     & 36.2$\pm$0.5 & 34.1$\pm$0.4 & 13.2$\pm$0.6 & 36.6$\pm$1.4 & 32.7$\pm$0.8 & 13.5$\pm$0.5 \\
CPT-Full    & 64.2$\pm$1.6 & 48.2$\pm$1.0 & 37.9$\pm$0.6 & 59.0$\pm$1.9 & 29.0$\pm$2.2 & 33.7$\pm$1.1 \\
CPT-CoocRM  & 66.4$\pm$1.9 & 42.7$\pm$2.0 & 34.6$\pm$1.9 & 52.1$\pm$1.7 & 24.1$\pm$2.0 & 34.0$\pm$2.2 \\
CPT-SubjRM  & 63.6$\pm$1.1 & 38.0$\pm$1.1 & 29.3$\pm$1.3 & 46.1$\pm$2.1 & 24.0$\pm$3.1 & 21.1$\pm$1.7 \\
CPT-SubjRP  & 64.5$\pm$1.3 & 43.7$\pm$0.9 & 32.9$\pm$0.9 & 51.1$\pm$0.7 & 30.3$\pm$3.1 & 24.6$\pm$1.7 \\
CPT-RandRM  & 65.0$\pm$0.9 & 47.8$\pm$1.5 & 37.0$\pm$1.2 & 56.9$\pm$1.5 & 31.1$\pm$2.3 & 32.2$\pm$1.5 \\
\bottomrule
\end{tabular}
\caption{Fact recall and cross-lingual transfer across model variants using a question answering template with random candidates. Random candidates are sampled uniformly from the same relation. \textsc{AccEN} and \textsc{AccFA} report top-1 accuracy in English and Persian; \textsc{Overall} is the fraction of triples answered correctly in both. \textsc{Transfer} is Persian accuracy conditioned on English correctness; \textsc{Non-Transfer} conditions on English incorrectness. \textsc{RankC} is the Spearman rank correlation between English and Persian candidate orderings.}
\label{tab:main-results-random}
\end{table*}

\end{document}

%% file: marco.tex
\begin{tikzpicture}[
  node distance = 1.5cm,
  every node/.style = {font=\large},
  mainnode/.style = {
    rectangle, rounded corners=4pt,
    draw, thick,
    minimum width=3.2cm, minimum height=0.9cm,
    align=center,
    fill=white,
  },
  notebox/.style = {
    rectangle, rounded corners=2pt,
    draw, dashed, thin,
    minimum width=3.2cm, minimum height=0.75cm,
    align=center, text=gray, font=\small % footnotesize
  },
  arr/.style = {-{Stealth[length=6pt]}, thick},
  redarr/.style = {-{Stealth[length=6pt]}, thick, red},
  circled/.style = {draw=red, fill=white, circle, inner sep=2pt, minimum width=1em, font=\sffamily\bfseries, text=red},
]

%% ── NODES ──

\node[mainnode] (enc) {English corpus};
\node[mainnode, right=6cm of enc] (pec) {Persian corpus};

\node[mainnode, above=2cm of enc] (enf) {English fact};
\node[mainnode, right=6cm of enf] (pef) {Persian fact};

\node[mainnode, below=0cm of $(enf.north)!0.5!(pef.north)$] (pfr) {Persian fact recall};

\node[mainnode, below=2cm of pfr] (sc) {Shallow cues};

%% ── NOTE BOXES ──

\node[notebox, above=0.3cm of enf] (enf-note) {\textbf{Einstein} died in \textbf{Princeton}.};
\node[notebox, above=0.3cm of pef] (pef-note) {
    \foreignlanguage{arabic}{%
        \textbf{اینشتین} در \textbf{پرینستون} درگذشت.%
    }
};
\node[notebox, above=0.3cm of pfr] (pfr-note) {
   FA: \foreignlanguage{arabic}{%
       محل فوت \textbf{اینشتین} \rule{1cm}{0.4pt} است.%
    }
    \\
    EN:\small\itshape{\textbf{Einstein} passed away in \rule{1cm}{0.4pt} .}
};

\node[notebox, below=0.3cm of sc]  (sc-note)  {FA: [\foreignlanguage{arabic}{ پرینستون، اولم، زوریخ، برلین، واشنگتن}] \\
{EN: \small\itshape[Princeton, Ulm, Zurich, Berlin, Washington]}};

%% ── ARCS BETWEEN NODES ──

% English corpus → English fact
\draw[arr] (enc) -- (enf);

% Persian corpus → Persian fact
\draw[arr] (pec) -- (pef);

% English fact → Persian fact recall (TRANSFER)
\draw[arr] (enf) -- (pfr);

% Persian fact → Persian fact recall (MEMO.)
\draw[arr] (pef) -- (pfr);

% Persian fact recall → Shallow cues
\draw[arr] (sc) -- (pfr);

\begin{pgfonlayer}{background}
    % English fact → Shallow cues (smooth left arc)
%    \draw[arr] ([xshift=-12pt]enc.south) -- ([xshift=-12pt]enf-note.south) to[out=270, in=180] (sc.west);
    \draw[arr] (enc) to (sc);

    % Persian fact → Shallow cues (smooth right arc)
%    \draw[arr] ([xshift=12pt]pec.south) -- ([xshift=12pt]pef-note.south) to[out=270, in=0] (sc.east);
    \draw[arr] (pec) to (sc);
\end{pgfonlayer}

%% ── CIRCLED LABELS ──

\node[circled, draw=black, text=black, right=0.3cm of enf] {a\vphantom{bq}};

% I1 between English fact and Persian fact
\node[circled, left=0.3cm of pef] {b\vphantom{bq}};

% I2 beside Shallow cues
\node[circled, above=0.7cm of sc] {c\vphantom{bq}};

\end{tikzpicture}

%% file: appendix_causality.tex
\section{Causal Effect Estimation of Factual Knowledge Transfer}
\label{app:causal-effect-estimation}

In this section, we motivate our experimental design from a causal perspective \citep{pearl2009causality}. Figure~\ref{fig:causal-graph} illustrates the assumed causal graph underlying our setup.

Let $\FactsEn$ denote whether the model has access to the relevant fact through English pretraining, and let $\FactsPers$ denote whether the fact is available through Persian continued pretraining. As discussed in (\S\ref{subsec:shallow-cues}), let $\ShallowCues$ represent shallow statistical cues that the model may exploit to make predictions without retrieving the underlying fact. Finally, let $\Prediction$ be a binary variable indicating whether the model predicts a fact correctly in Persian.

For simplicity, we treat $\FactsEn$, $\FactsPers$, $\ShallowCues$, and $\Prediction$ as binary variables. For example, $\FactsPers = 0$ corresponds to a setting where the relevant fact has been removed from the Persian training data, while $\FactsPers = 1$ denotes that the fact remains available.

Our goal is to estimate the extent to which correct Persian predictions can be causally attributed to factual knowledge acquired from English. Ideally, one would estimate the average causal effect of English factual knowledge by directly intervening on $\FactsEn$:
\begin{equation}
\E[\Prediction \mid \causalDo(\FactsEn = 1)]
-
\E[\Prediction \mid \causalDo(\FactsEn = 0)]
\end{equation}

\input{dag}

However, such interventions are infeasible in our setting because the models are initialized from an English-pretrained checkpoint. In particular, we never observe the counterfactual model that would have been trained identically but without exposure to the relevant English facts.

Instead, we study a controlled causal effect under conditions where alternative sources of information are removed. Specifically, we consider the quantity
\begin{equation}
\label{eq:cace}
\begin{aligned}
&\cace
\coloneq \\
&\ \ \E[\Prediction \mid \causalDo(\FactsEn = 1), \causalDo(\FactsPers = 0), \causalDo(\ShallowCues = 0)] \\
&-
\E[\Prediction \mid \causalDo(\FactsEn = 0), \causalDo(\FactsPers = 0), \causalDo(\ShallowCues = 0)]
\end{aligned}
\end{equation}

This quantity measures the causal contribution of English factual knowledge when neither Persian factual exposure nor shallow heuristics can support prediction. Intuitively, under these interventions, correct predictions should primarily reflect cross-lingual factual transfer from English.

Importantly, this controlled effect differs from the unconditional average causal effect above. If the relevant fact remains directly available in Persian training data, adding English factual knowledge may have little effect on prediction accuracy because the model can already solve the task from Persian evidence alone. Our controlled intervention therefore isolates the model's ability to transfer knowledge across languages rather than its overall factual recall performance.

We approximate the intervention $\causalDo(\FactsPers = 0)$ by systematically removing facts from the Persian training corpus, as described in (\S\ref{subsec:removal}). While we cannot directly intervene on $\ShallowCues$, we construct a hard-candidate evaluation setting designed to minimize the predictive utility of shallow heuristics. We assume that the probability of success in this hard setting is approximately equal to the probability of success in the random setting had we removed the shallow cues through an intervention
\begin{equation}
\label{eq:do-fact-en-hard}
\begin{aligned}
&\E[\Prediction_{\text{hard}} \mid \causalDo(\FactsEn = 1), \causalDo(\FactsPers = 0)]
\\
&\approx
\E[\Prediction \mid \causalDo(\FactsEn = 1), \causalDo(\FactsPers = 0), \causalDo(\ShallowCues = 0)],
\end{aligned}
\end{equation}
\noindent i.e.\ within this evaluation setting, the remaining contribution of shallow cues is negligible, approximating the intervention $\causalDo(\ShallowCues = 0)$.

Under these interventions, we can relate the first term in $\cace$ to an observable quantity:
\begin{equation}
\label{eq:do-fact-en}
\begin{aligned}
&\E[\Prediction \mid \causalDo(\FactsEn = 1), \causalDo(\FactsPers = 0), \causalDo(\ShallowCues = 0)]
\\
&=
\E[\Prediction \mid \FactsEn = 1, \causalDo(\FactsPers = 0), \causalDo(\ShallowCues = 0)] \\
&\approx
\E[\Prediction_{\text{hard}} \mid \FactsEn = 1, \causalDo(\FactsPers = 0)]
\end{aligned}
\end{equation}

This corresponds to the setting evaluated by our intervention experiments: the model retains access to English-acquired knowledge, while Persian factual evidence and shallow heuristics are minimized.

Although we cannot directly observe the counterfactual intervention $\causalDo(\FactsEn = 0)$, under our controlled setup we expect performance without informative English facts to fall approximately to chance level:
\begin{equation}
\label{eq:no-fact-en}
\begin{aligned}
    & \E[\Prediction \mid \causalDo(\FactsEn = 0), \causalDo(\FactsPers = 0), \causalDo(\ShallowCues = 0)]\\
    & \approx
    \Prediction_{\mathrm{chance}}
\end{aligned}
\end{equation}
\noindent with $\Prediction_{\mathrm{chance}} = 0.2$ because the candidate set contains five options.

Together, Equations~\ref{eq:do-fact-en} and \ref{eq:no-fact-en} motivate our experimental design: by removing Persian factual evidence and minimizing shallow cues, successful prediction above chance level provides evidence that the model is relying on factual knowledge transferred from English.

%% file: dag.tex
\begin{figure}[t]
    \centering
    \begin{tikzpicture}[
      scale=0.8,
      transform shape,
      every node/.style={align=center, font=\small},
      arr/.style={->, thick, >=Stealth}
    ]

    \node (eng) at (0,2) {English\\corpus};
    \node (per) at (0,0) {Persian\\corpus};

    \node (fen) at (3,2) {Facts\\Eng.};
    \node (fpe) at (3,0) {Facts\\Pers.};

    \node (cues) at (6,3) {Shallow Cues};
    \node (pred) at (6,1) {Persian Factual \\Prediction};

    \draw[arr] (eng) -- (fen);
    \draw[arr] (per) -- (fpe);

    \draw[arr] (fen) -- (pred);
    \draw[arr] (fpe) -- (pred);

    \draw[arr] (cues) -- (pred);

    \draw[arr, bend left=15] (eng) to (cues);
    \draw[arr] (per) to[out=120, in=-155] (-0.6, 2.7) to[out=25, in=160] (cues);

    \end{tikzpicture}
    \caption{Causal graph illustrating the relationship between corpora, extracted facts, shallow cues, and factual prediction in Persian.}
    \label{fig:causal-graph}
\end{figure}